# A Machine learning approach for rapid disaster response based on multi-modal data.

Machine learning for shelter needs assessment

A Machine learning approach for rapid disaster response based on multi-modal data. The case of housing & shelter needs.

Karla Saldana Ochoa

University of Florida, College of Design, Construction & Planning, Faculty of Architecture,
ksaldanaochoa@ufl.edu

Tina Comes

TU Delft, Faculty Technology, Policy & Management, t.comes@tudelft.nl

Along with climate change, more frequent extreme events, such as flooding and tropical cyclones, threaten the livelihoods and wellbeing of poor and vulnerable populations. One of the most immediate needs of people affected by a disaster is finding shelter. While the proliferation of data on disasters is already helping to save lives, identifying damages in buildings, assessing shelter needs, and finding appropriate places to establish emergency shelters or settlements require a wide range of data to be combined rapidly. To address this gap and make a headway in comprehensive assessments, this paper proposes a machine learning workflow that aims to fuse and rapidly analyse multimodal data. This workflow is built around open and online data to ensure scalability and broad accessibility. Based on a database of 19 characteristics for more than 200 disasters worldwide, a fusion approach at the decision level was used. This technique allows the collected multimodal data to share a common semantic space that facilitates the prediction of individual variables. Each fused numerical vector was fed into an unsupervised clustering algorithm called Self-Organizing-Maps (SOM). The trained SOM serves as a predictor for future cases, allowing predicting consequences such as total deaths, total people affected, and total damage, and provides specific recommendations for assessments in the shelter and housing sector. To achieve such prediction, a satellite image from before the disaster and the geographic and demographic conditions are shown to the trained model, which achieved a prediction accuracy of 62%.

# 1 INTRODUCTION

Each year, natural disasters destroy thousands of homes. In 2019, almost 2,000 natural hazards provoked 24.9 million new displacements across 140 countries; three times the number of displacements caused by conflict [1]. Data provide tremendous opportunities for earlier warnings and faster aid that is better tailored to the needs of the population [1]. One of the key challenges in disasters is combining the plethora of available information, ranging from social media data to satellite imagery, and turning it into concrete and actionable insight [3]. This is especially true for humanitarian shelter [4]. Such shelter responses can fail to achieve adequate solutions when faced with spatial and constructive constraints. By learning from past experiences, (local) decision-makers with limited prior experience can approach a new event with information that includes architectural knowledge in a clear and organized manner. This requires multi-modal data on disaster impact, location-based risk and spatial conditions

Research has shown that using multimodal data for machine learning leads to more robust inference than learning from a single modality [5,6,7,8]. There are two approaches to multimodal data fusion. The first is feature-level fusion, also known as early fusion. The second is decision-level fusion or late fusion, which fuses multiple modalities into one semantic space [5]. The latter will be used in this research, as the modalities used in a fusion process may provide complementary or contradictory information, so it is necessary to know which modalities contribute or not in the final inference. Within this line of research, several probabilistic AI models have been proposed to address the multimodal scenario using decision-level (or late) fusion in tasks such as audiovisual analysis [7,8], object detection [9,10], speech processing [11], and scene recognition [12,13]. Particularly for disaster response, one branch of research uses this approach through multimodal fusion models with social media data, taking into account posts that include images and text during a natural disaster to create automatic classifiers [14,15,16,17].

Although multimodal data fusion for disaster response has been applied to social media data, a research gap is including other sources, such as satellite imagery or statistical data, to capture the spatial context or exposure. Further, social media is increasingly prone to misinformation. By broadening the knowledge base, inferences can be generated based on information about past natural disasters. These can be used to make decisions in current situations that may have similarities to past events.

The objective of this paper is to create a proof of concept for an instrument that can guide decision-makers at the operational level in the shelter and housing sector. This paper's contribution is twofold. First, it presents a methodology for multimodal fusion at the decision level to characterize a natural disaster; images, text, and statistical data. Second, it offers a machine learning model trained with freely available data that supports decision-makers by predicting the degree of damage in terms of casualties, affected population, and material losses, including recommendations for the shelter and housing sector.

# 2 METHODOLOGY

A machine learning process consisted of three stages: teaching, learning, and inference, which are presented in this section. The first, teaching, focused on the data that served as training data. Here the data handling process is an essential task. The second stage was learning; which focused on the architecture of the ML algorithm that would learn the training data. The third, inference, focused on the validation of the obtained results.

### 2.1 Teaching

**Data.** We used three types of data sources about (i) disaster characteristics; (ii) location-based risk; and (iii) spatial conditions. To ensure scalability and low cost of our approach, we only used open and online data. (i) For the disaster characteristics, we used EM-DAT (https://www.emdat.be/) as a basis, and enriched it with additional values regarding demographic and geographic conditions for disasters from 2010-2019. We consider seven attributes to describe each natural disaster: location, time, type of disaster, magnitude of the disaster, total number of deaths, total number of people affected, and total damages (US$). Additionally, we record the epicenter of the disaster in (lat/lon) and date of occurrence. Out of the total recorded disasters by EM-DAT, the 202 are selected that provide the complete dataset. (ii) For location-based risk, we include the World Bank's Global Urban Risk Index [18] and the INFORM index proposed by the European Commission, which measures the performance in humanitarian risk management. In addition, we use UNdata (https://data.un.org/) for elevation data, climate types, natural resources, employment to population ratio, construction value-added, total renewable electricity, population, and area of the three nearest administrative divisions. For climatic conditions, we added Sea Temperature Conditions (STC) at the time of the disaster. (iii) To capture the spatial conditions and extract urban features such as road infrastructure, rivers or land use that drive shelter and housing decisions, we used satellite imagery from Google Maps. The specifics of the processing of the satellite imagery are provided in Appendix A.1. The data collection results in a database including 202 natural disasters with 19 attributes or independent features. A full summary is provided in Appendix A.2

The next steps of preprocessing, dimensionality reduction and normalization are performed to support the hyper-parameter fitting process and ensure the quality of the final inference. In the case of this experiment, it guarantees the successful late function of the different data modalities for the prediction.

**Preprocessing.** Each data modality was preprocessed to remove outliers (extreme high/low values) and extract its numerical values. Alphanumerical and alphabetical data was converted into numerical values using the one-hot encoding method. This method defines a dictionary of all words included in the data and assigns a number to each that encodes each word as an integer. Appendix A.3 shows five example entries.

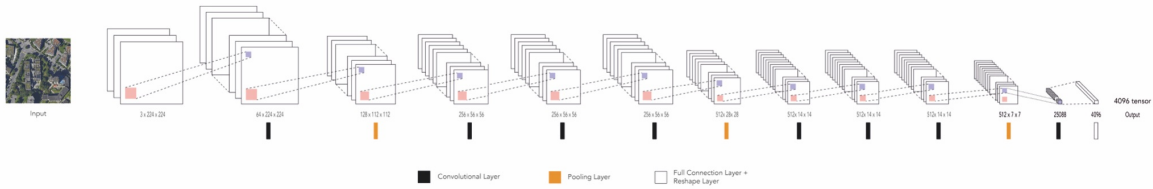

Figure 1: Feature Extraction part of the image identification VGG-Net Model.

To extract numerical features from the satellite images, a self-referential process was implemented via an unsupervised feature extraction algorithm to translate each image (per epicenter, we collected 2500 images patches that together capture the area desired) into an n-dimensional numerical vector. This process is unsupervised since the satellite images are provided as a set of unlabeled examples. To perform feature extraction, we first built an AI feature extractor (FE) model, which was made by cutting the last layers of a trained VGG-16 network (stack of convolutional layers, followed by fully connected layers) [19] and using the convolutional layers (feature extraction part), see Figure 1. We trained it with a large set of images (ILSVRC2012, 2012) (about 10 million images and 10,000 classes) used for the ILSVRC-2012 [20] challenge. However, the model can still learn from satellite imagery that was not included in its initial training data. To retrain the FE model, we sampled 90 images (10,000x10,000 pixels), corresponding to about 40% of the dataset, resized to 224×224 pixels in 3 RGB bands resulting in 4500 patches [21]. After the FE was trained, we extracted the features from the whole dataset, resulting in a list of 10100 numerical vectors of 4096 dimensions representing 202 natural disasters.

**Dimensionality reduction.** There are 3 data classes, whose dimensions are to be reduced: Climate types (text, dim=30), natural resources (text, dim=80), satellite images (image, dim=4096). We applied the nonlinear dimensionality reduction algorithm t-Distributed Stochastic Neighbour Embedding (t-SNE). The t-SNE algorithm can capture much of the local structure of high-dimensional data while revealing the global structure [22]. Within t-SNE we set a Perplexity of 100 as recommended by [23] to reduce each n-dimensional vector list to a 1-dimensional vector.

**Normalization.** Logarithmic Transformation (LT) reduces the positive and negative skewness of the data. The LT considers the absolute difference between values of similar performance and reduces the gap between values. This transformation differentiates more clearly the slight differences in all performance ranges and improves the interpretation of differences between values at opposite extremes. Figure 2 show the histograms after LT.

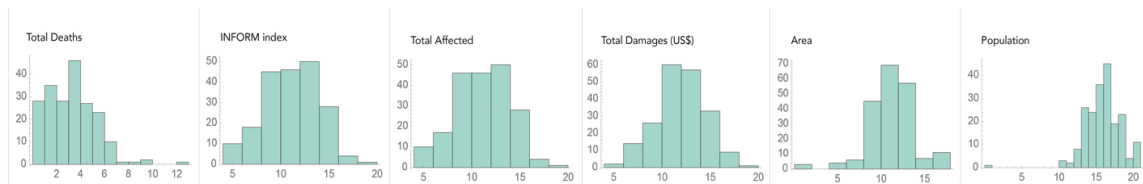

Figure 2: Histogram of the attributes after applying the Logarithm Transformation (LT)

**Standardization.** Rescales the values of each attribute to a zero mean and a sample variance equal to one. This method allows us to work in conjunction with variables that represent different modalities. To standardize, we calculate the standard deviation (σ) and then we calculate the standardization (μ) μ=(x-mean(x))/σ. After standardizing all attributes, they are ready to be fused to a vector with 19 dimensions describing each natural disaster. Each attribute contributes to the overall representation of the natural disaster.

## 2.2  Learning: Clustering Algorithm

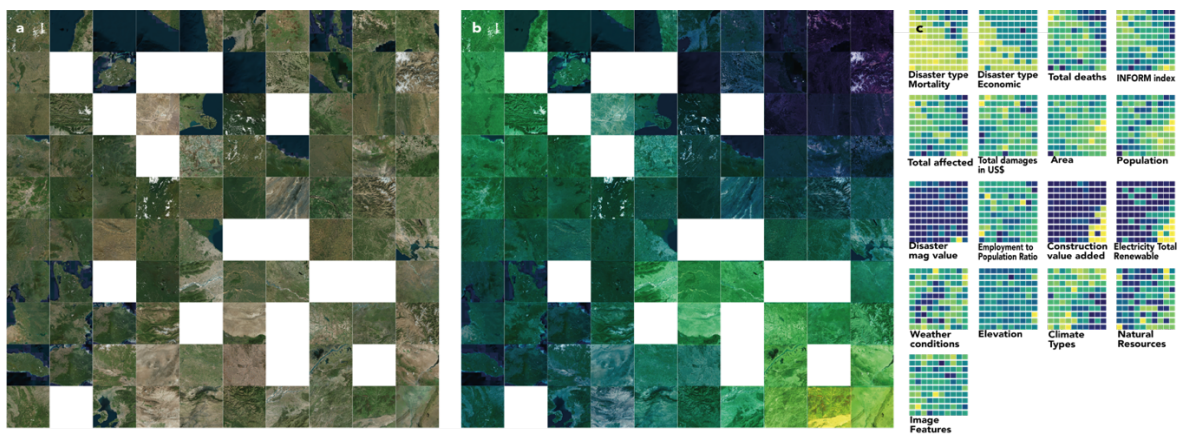

Figure 3. a) grid SOM displaying the SOM BMUS with satellite images; b) grid SOM displaying the SOM BMUS in color spectrum; c) BMUS color spectrum for each feature after training. For visualization purposes, a color assigned to the weight value (n-dimensional vector) from green to yellow is showing the consistency in the clustering.

To explore the relationships of the 202 disasters, we use Self Organizing Map (SOM), an unsupervised machine learning clustering algorithm [23]. We chose unsupervised machine learning as it allows us to explore the characteristics and features of disasters that lead to similar knowledge rather than making a-priori assumptions. By not predefining groups or classes but letting the data define those classes, we avoid bias in the inference. Further, SOM provide an analogy to maps, which are frequently used decision support tools in disaster management, whereby proximity represents similarity. We select a 10X10 2D gridded SOM. The input layer of the SOM consists of 19-dimensional vectors obtained in the previous steps. The training procedure involved one million epochs until the response layer evolved to a stable configuration. The output layer of the SOM is visualized as a similarity landscape, where each natural disaster is positioned according to the similarity of its weight value concerning its neighbors. Each node of the SOM or Best Matching Unit (BMU) represents the average of the original n-dimensional observed data that, after iteration, belongs to that node. Fig. 3 (left) shows the images (in this case, satellite images assigned for each node) of the natural disasters that have the closest Euclidean distance to their corresponding BMU value. A trained SOM can further visualize each feature corresponding to each BMU (Fig. 3. right), allowing decision makers to explore which feature contributed in which way to the overall result, or identifying which features influence the prediction the most. As such, SOM are transparent avoid many of the explainability issues present in Artificial Neural Networks. Apart from visualization, a trained SOM can be used as a predictor for missing features in new data input.

### 2.3 Inference: Validation Data

To validate the prediction, we focus on data from 2020 on natural disasters (earthquakes, storms, floods, and landslides), which was not included in the data used for prediction. We chose 20 natural catastrophes from 2020 as Validation Data (VD). Each natural disaster in the VD is encoded, normalized, and fused, as explained in the previous sections. The values to be predicted are casualties, number of people affected, and economic damage (US$). As this step is to provide decision-makers with a severity assessment of the disaster that should allow them to assess the required resources, we follow the classification of EM-DAT. Therefore, the prediction will not be an exact number but will assign a range of impact, such as low, medium, and high, to each attribute, which are the following:

Table 1: Assessment categories

| Degree | Casualties | People Affected | Economic Damage [US$] |
|---|---|---|---|
| Low | 1-50 | 1-10.000 | 1-100.000 |
| Mid | 51-5.000 | 10.0001- 5'000.000 | 100.001-5'000.000 |
| High | 5.001-30.000 | 5'000.0001 - 30'000.000 | 5'000.001-60'000.000 |

In addition to predicting consequences, the trained SOM will present a list of suggestions tailored to the shelter sector to give possible solutions that answer the question of what should be done. This will be based on the learning from previous experiences. Such an approach helps decision makers by giving them the liberty to decide how they are going to act.

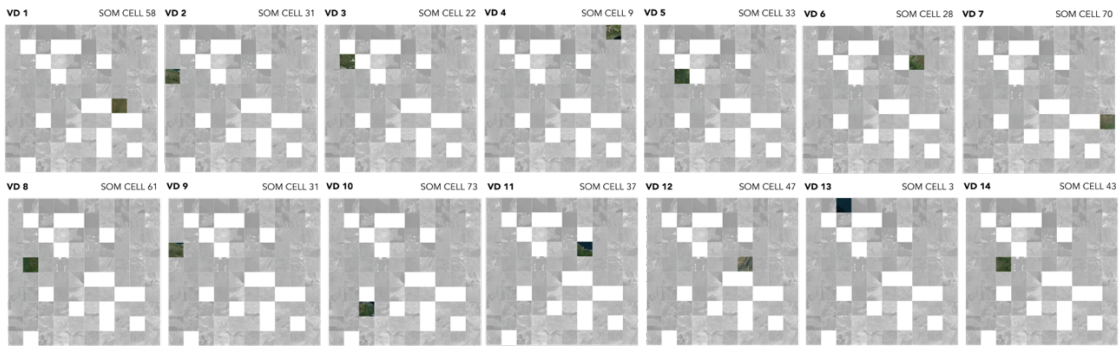

Figure 4. SOM cells activated with each query

After feeding the data to the trained SOM, each of the 20 natural disasters from the VD was matched to a node or a BMU (Figure 4). Hence, a BMU of the trained SOM represents a single historical disaster that best represents the cluster in which a new data point belongs. The predictions are thus the impacts of that historical disaster, and the "recommendations" are those produced after that historical disaster. To validate the performance of the trained SOM, we compare the prediction (total deaths, total affected, and total damage (US$)) with the ground truth labels. Appendix A.4 shows a table with the result of the comparison, showing that of the 42 ground truth data points, 26 were correctly labeled, 7 were over predicted, and 9 were under predicted with only one degree more or one degree less, and the highest accuracy in the category number of affected people (78% correct). In total, we arrived at a prediction of 62%. However, in the face of a natural disaster, following the precautionary principle, overprediction – especially if it is with respect to only one of the categories – is more acceptable than underprediction. If we add the correct predictions with the over predictions, we arrive at a model that has 83% non-under prediction.

## 3 DISCUSSION: RESULTS

Predicting the impact of a natural disaster on a community in terms of total number of deaths, total number of people affected, and total damages ($US) allows decision-makers to articulate early decisions, influencing the outcome of the following operations in the disaster response. These data (expected impact) help with having situational awareness in general terms (Table 2). However, one requires specific information to facilitate decisions in individual clusters of the response, such as the shelter and housing sector.

Table 2. Description of the validation data and prediction in terms of total deaths, total affected, and total damages.

| BMU | Country | Year | Type Disaster | Deaths | Affected | Damages |
|---|---|---|---|---|---|---|
| 58 | Brazil | 2020 | Flood | mid | mid | mid |
| 31 | Indonesia | 2019 | Flood | mid | mid | mid |
| 22 | Nepal | 2017 | Flood | mid | mid | mid |
| 9 | Italy | 2014 | Flood | low | low | mid |
| 33 | Vietnam | 2013 | Flood | low | mid | low |
| 28 | Ukraine | 2010 | Flood | low | mid | low |
| 70 | China | 2010 | Earthquake | low | mid | low |
| 61 | Philippines | 2011 | Flood | low | mid | low |

| 31 | Indonesia | 2019 | Flood | mid | mid | mid |
| 73 | Fiji | 2012 | Flood | low | low | low |
| 37 | Indonesia | 2016 | Earthquake | mid | mid | low |
| 47 | Pakistan | 2013 | Earthquake | mid | mid | low |
| 3 | Mexico | 2017 | Earthquake | mid | mid | mid |
| 43 | Kenya | 2012 | Flood | mid | mid | low |
| 30 | Myanmar | 2016 | Earthquake | low | low | low |
| 26 | Papua New Guinea | 2018 | Earthquake | mid | mid | low |
| 10 | Greece | 2016 | Flood | low | low | low |
| 47 | Pakistan | 2013 | Earthquake | mid | mid | low |
| 49 | India | 2013 | Earthquake | low | mid | mid |
| 29 | Myanmar | 2012 | Earthquake | low | low | low |

To achieve this inclusion of specific knowledge, we will focus on the information provided by The Global Shelter Cluster (http://www.shelterprojects.org). Over the past 20 years, The Global Shelter Cluster has published lessons learned and recommendations for natural disasters and listed recommendations for the construction of shelters and temporary houses. These recommendations can be translated into a list of potential assessments that will strengthen the response and identify assessments that may reduce the effectiveness of the response. Such information will be searched for and assigned to each of the representative catastrophes in each SOM cell or BMU from the trained SOM. Therefore, when a VD point is assigned to a specific SOM cell or BMU, such trained SOM will not only predict the impact of the disaster but will also show a list of recommendations for shelter and temporary housing sector after a natural disaster (Figure 5) that are based on previous experiences.

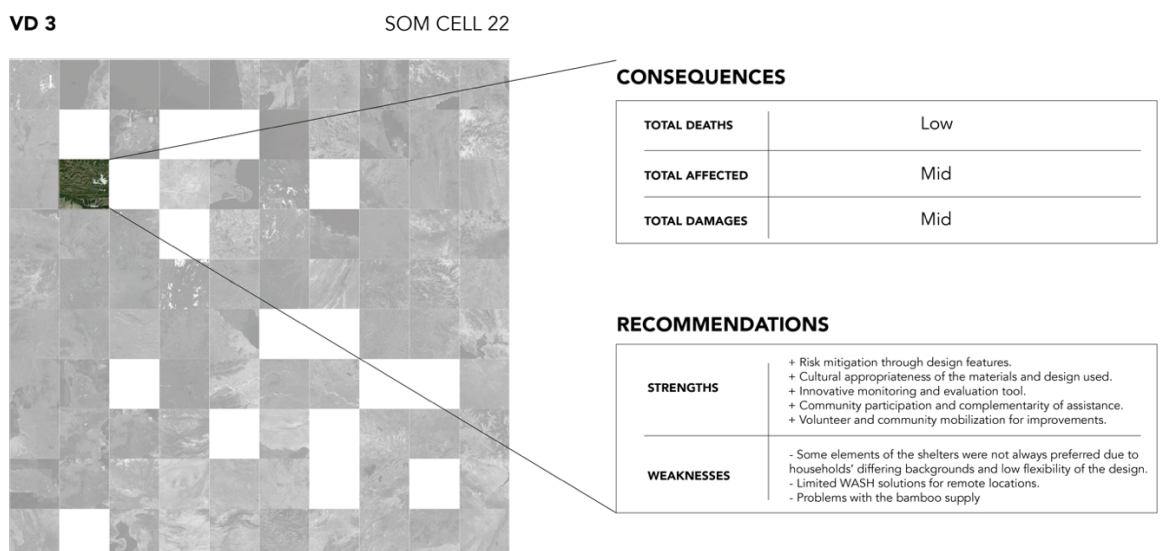

Figure 5. An example of how a Personalized Event Protocol, here the VD 3 activated SOM Cell 22, the instrument predicted the consequences and added specific recommendations for the assessment that deals with shelter and temporary housing.

## 4 CONCLUSIONS

In this paper, we presented a Machine Learning workflow that turns open and online multi-modal data into results to support rapid decisions after disasters. We showcase our approach by applying it to the case of shelter needs after natural disaster response. We first established a database of 202 natural disasters characterized by different data modalities that refer to specific characteristics of the natural disaster and geographic and demographic constraints. Subsequently, our approach uses different approaches for feature extraction from each data modality (text, satellite images, statistical data, indexes) and then use the Standardization method for late fusion. After transforming all modalities to numerical representations, we normalized their data to ensure that late fusion did not prioritize some features over others and that all influenced to a similar degree to the final inference. These fused vectors were input into an unsupervised machine learning algorithm SOM, which clustered the disasters based on the similarities in their numerical features. To validate the experiment, data on natural disasters from 2020 were used, obtaining an accuracy of 63%. The prediction includes specific information to guide decisions on the shelter and housing sector. With such information, we make a headway in improving situational awareness by rapidly providing the much-needed assessments.

An analysis of the trained SOM (Fig. 3) shows that some characteristics had a greater influence over the final clustering: (1) Casualties; (2) Total affected; (3) Total damage (US$); (4) Global urban risk index economic; (5) Global urban risk index mortality; (6) Sector elevation; (7) Sector construction value added; (8) Population of the three nearest administrative divisions; (9) Area of the three nearest administrative divisions'; (10) Urban features from satellite images. These characteristics will be considered in further research.

Our approach allows us to analyse a natural disaster as a part of a set of disasters that may show similarities according to the different characteristics considered. Therefore, this approach allows for training, and cross-organizational or cross-contextual learning, which is often missing in humanitarian organizations [24]. By including more data in the database, more specific predictions of different consequences can be made [25]. In the present case, general attributes of the population and the sector were considered. While the attributes selected do not fully capture the complexity of decisions in natural disasters, our approach provides a starting point and can be consider as an initial prove of concept to expand, also beyond the sector of shelter & housing. Future research can include social media sources such as to capture sentiment or humanitarian needs, or more specific data on humanitarian organizations' capacities or responses as well as capacity and infrastructural information.

## ACKNOWLEDGMENTS

The authors would like to thank the Reviewers and Editor for their helpful comments and constructive suggestions.

## A    APPENDICES

### A.1    Satellite Imagery

A data modality that is rarely correlated with statistical data and urban indices on natural disasters is the spatial conditions of a sector. Such spatial conditions and urban features range from recognizing transportation networks (roads, rail); major infrastructure and characteristic features (rivers, coasts) as well as land use. As good connections to transportation lines as well as limiting exposure to potential subsequent disasters (e.g., landslides after earthquakes) is vital for post-disaster shelter locations, including such features is an important step to support decisions. One way to include this characteristic would be through the use of satellite imagery. Today there are online services such as Google Maps, where satellite images of most places on the planet are available. Therefore, this source can be used to collect free satellite images of almost any place on Earth. It is important to note that our database is composed of data that is open and online, thus ensuring its scalability.

To collect the satellite images of each catastrophe, we used the Google Maps API to define a grid of coordinates covering a natural disaster epicenter. In our case, such a grid was 10x10 kilometers (202 grids of 10x10 kilometers, one for each natural disaster) with a zoom level (Z) of 18, where each point of the grid is separated by 200 meters, which means that for each natural disaster, 2500 satellite images were collected each covering 200x200m without overlapping (Figure 6). With a total of 505000 patches of satellite images.

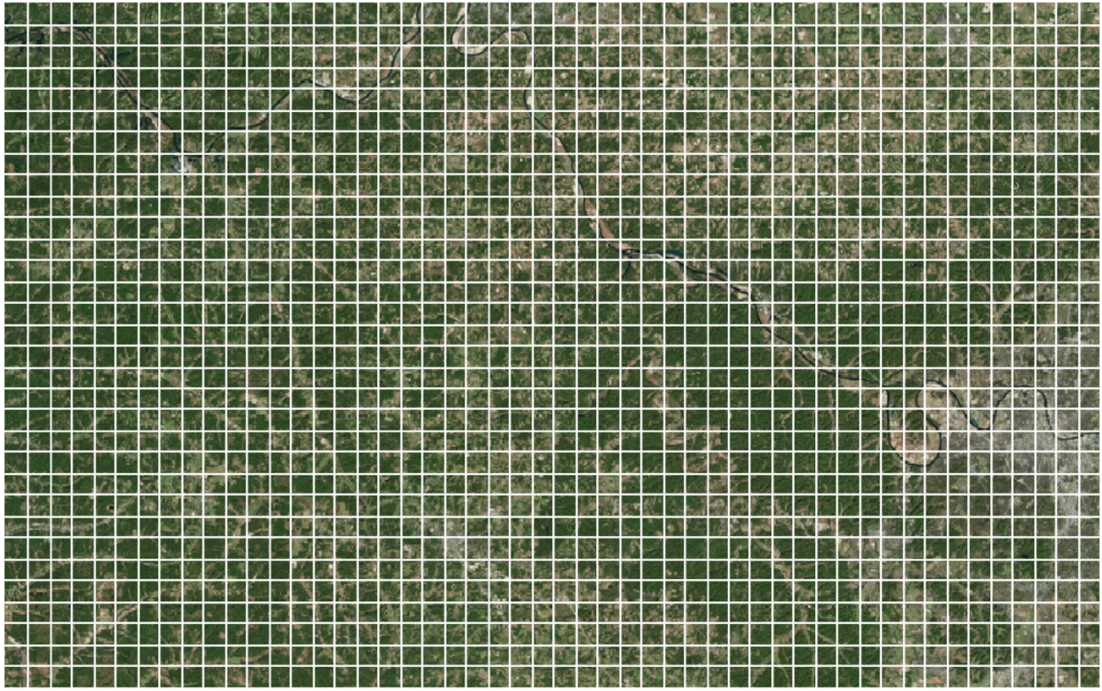

Figure 6. An example of the patches for each natural disaster, which later was joined to create a whole satellite image.

After collecting the satellite images, it is necessary to correct the scale to make them comparable. All satellite images are corrected for latitude deformation using equation 3, which calculates the terrain resolution (tr) in meters per pixel. Where 6'378.137 refers to the radius of the Earth (RE) in meters, which helps define the earth's circumference (ec) in equation 1. Considering a map width (mw) calculated in equation 2.

$$ec = 2\pi \times RE \tag{1}$$

$$mw = 256 \times 2Z \tag{2}$$

$$tr = Cos\left(lat \times \frac{\pi}{180}\right) \times \frac{ec}{mw} \tag{3}$$

## A.2 Data Collection and sources

Table A.1. Description of the attributes contained within the MAP-DIS database

| Class | Description | Source | Modality |
|---|---|---|---|
| 1 Location | Coordinates of the epicenter of the disaster | EM-DAT | number |
| 2 Time | Day, month and year of the occurrence of the disaster | EM-DAT | number |
| 3 Disaster Magnitude | Value of magnitude depending type of disasters | EM-DAT | number |
| 4 Total Deaths | Sum of death and missing. | EM-DAT | number |
| 5 Total Affected | Sum of injured, homeless, and affected. | EM-DAT | number |
| 6 Total Damages (US$) | The amount of damage to property, crops, and livestock. The value of estimated damage is given in US$ ('000). | EM-DAT | number |
| 7 Urban risk index Economy | Encoded based on a Global Urban Risk Index on mortality risk and economic lost | World Bank | number |
| 8 Urban risk index Mortality | Encoded based on a Global Urban Risk Index on mortality risk and economic lost | World Bank | number |
| 9 INFORM index | Generalized risk of a crisis based on structural conditions | European Commission | number |
| 10 Elevation | Elevation point of the epicenter | UN Stats | number |
| 11 Climate Types | A list of type of climates found in the region | UN Stats | text |
| 12 Natural Resources | A list of natural resources found in the region | UN Stats | text |
| 13 Employment to Population Ratio | Is the proportion of a country's working-age population that is employed. This indicator is expressed as a percentage | UN Stats | number |
| 14 Construction Value Added | Is the output of the sector after adding up all outputs and subtracting intermediate inputs. | UN Stats | number |
| 15 Electricity Total Renewables | Renewable electricity output (% of total electricity output) | UN Stats | number |
| 16 Area | Area of the three closest cities from the epicenter | UN Stats | number |
| 17 Population | Population of the three closest cities from the epicenter | UN Stats | number |
| 18 Temperature Conditions | Sea Surface temperature (SST) | UN Stats | number |
| 19 Satellite Image | Satellite imagery from the site of the natural disaster | Google Earth | image |

## A.3 Preprocessing of alpha-numerical and alphabetical data

Table A.2. Five entries with their initial text description and then transformed to a numerical value.

| # | Text Description | Numerical Value |
|---|---|---|
| 1 | cold semi-arid cold desert temperate oceanic subpolar oceanic warm-summer Mediterranean cold-summer Mediterranean polar tundra | 0, 2, 1, 0, 1, 0, 0, 0, 0,0, 0, 0, 0, 2, 0, 0, 0 ,2,1, 0, 0, 1, 0, 0, 1, 0, 1, 0, 1, 1 |
| 2 | dry-winter tropical savanna hot semi-arid hot desert | 0, 0, 0, 0, 1, 0, 1, 0, 0, 2, 0, 0, 0, 0, 0, 0, 0, 0, 0, 0, 1, 1, 0, 0, 0, 0, 0, 1, 0, 0 |
| 3 | tropical monsoon dry-summer tropical savanna dry-winter tropical savanna hot semi-arid cold semi-arid hot desert cold desert humid subtropical temperate oceanic hot-summer Mediterranean warm-summer Mediterranean monsoon-influenced humid subtropical | 0, 2, 0, 0, 2, 1, 1, 0, 0, 2, 1, 2, 0, 2, 0, 1, 1, 1, 0, 0, 2, 2, 0, 0, 0, 2, 1, 3, 0, 1 |
| 4 | tropical monsoon dry-winter tropical savanna cold semi-arid cold desert humid subtropical monsoon-influenced humid subtropical subtropical highland cold subtropical highland hot-summer humid continental warm-summer humid continental subartic monsoon-influenced hot-summer humid continental monsoon-influenced warm-summer humid continental monsoon-influenced subarctic polar tundra | 0, 3, 0, 4, 1, 0, 1, 0, 2,0, 2, 6, 0, 0, 0, 1, 4, 0, 1, 0, 1, 1, 1, 1, 0, 4, 0, 2, 1, 2 |
| 5 | tropical rainforest tropical monsoon dry-summer tropical savanna dry-winter tropical savanna hot semi-arid temperate oceanic warm-summer Mediterranean subtropical highland | 0, 0, 0, 0, 0, 1, 1, 0, 1, 1, 0, 0, 0, 1, 0, 1, 0, 1, 0, 1, 2, 1, 0, 0, 0, 1, 1, 4, 0, 1 |

## A.4 Result of the prediction

We will focus on data from 2020 on natural disasters (earthquakes, storms, floods, and landslides) that share location and time of occurrence. These disasters conform a Validation Data (VD), will be encoded, normalized, and fused, as explained in the previously. Of all the natural disasters that occurred in 2020, 20 were selected. This selection of disasters covers the following countries: Brazil, Vietnam, Madagascar, Puerto Rico, Uganda, China, Greece, Croatia, Indonesia, Iran, Mexico, Malaysia, Philippines, and Turkey.

To validate the performance of the proposed workflow, we can compare the prediction (total deaths, total affected, and total damage (US$)) with the ground truth labels from the VD. Some of the natural disasters had only one or two of the attributes. For example, a disaster in Madagascar on January 17, 2020, had values on total deaths and total affected people, but not on total damage. Hence, of the 60 ground truth data from the VD, only 42 values were suitable for comparison.

Table A.3. Three tables comparing the ground truth with the prediction of the SOM

| total deaths | |
| --- | --- |
| ground truth | prediction |
| mid | mid |
| low | mid |
| low | mid |
| no info | low |
| low | low |
| low | low |
| low | low |
| low | low |
| low | mid |
| low | low |
| no info | mid |
| no info | mid |
| low | mid |
| no info | mid |
| low | low |
| low | mid |
| low | low |
| low | mid |
| no info | low |
| mid | low |

| no affected | |
| --- | --- |
| ground truth | prediction |
| mid | mid |
| mid | low |
| mid | mid |
| low | low |
| mid | mid |
| mid | mid |
| mid | low |
| mid | mid |
| mid | mid |
| low | low |
| mid | mid |
| mid | mid |
| mid | mid |
| mid | low |
| low | low |
| mid | mid |
| low | low |
| mid | low |
| mid | low |
| low | low |

| total damages | |
| --- | --- |
| ground truth | prediction |
| mid | mid |
| low | mid |
| no info | mid |
| no info | mid |
| no info | low |
| mid | low |
| low | low |
| no info | low |
| mid | mid |
| mid | low |
| no info | low |
| no info | low |
| no info | mid |
| no info | low |
| no info | low |
| no info | low |
| no info | low |
| no info | low |
| no info | mid |
| mid | mid |

| | | | |
| --- | --- | --- | --- |
| correct | 26 | 62 | % |
| incorrect | 7 | 17 | % |
| overprediction | 9 | 21 | % |

| | | |
| --- | --- | --- |
| total | 42 | |
| correct + overprediction | 83 | % |
| error | 17 | % |

13